\documentclass[sigconf,nonacm]{acmart}
\setcopyright{none}
\AtBeginDocument{%
  }

\usepackage{amsmath}
\usepackage{mathtools}
\usepackage{amsthm}
\usepackage{comment}
\usepackage{multirow}
\usepackage{arydshln}
\usepackage{algorithm}
\usepackage{algpseudocode}
\usepackage[capitalize]{cleveref}
\usepackage{enumitem}

\begin{document}

\title{WRF4CIR: Weight-Regularized Fine-Tuning Network for Composed Image Retrieval}

\author{Yizhuo Xu$^1$, Chaojian Yu$^1$, Yuanjie Shao$^1$, Tongliang Liu$^2$, Qinmu Peng$^1$, Xinge You$^1$ \\
$^1$Huazhong University of Science and Technology\\
$^2$University of Sydney\\
\texttt{\{xuyizhuo1, chaojianyu, shaoyuanjie, pengqinmu, youxg\}@hust.edu.cn} \\ \texttt{tongliang.liu@sydney.edu.au}}

\renewcommand{\shortauthors}{Yizhuo Xu, Chaojian Yu, Yuanjie Shao, Tongliang Liu, Qinmu Peng, Xinge You}

\begin{abstract}
Composed Image Retrieval (CIR) task aims to retrieve target images based on reference images and modification texts. Current CIR methods primarily rely on fine-tuning vision-language pre-trained models. However, we find that these approaches commonly suffer from severe overfitting, posing challenges for CIR with limited triplet data. To better understand this issue, we present a systematic study of overfitting in VLP-based CIR, revealing a significant and previously overlooked generalization gap across different models and datasets. Motivated by these findings, we introduce WRF4CIR, a Weight-Regularized Fine-tuning network for CIR. Specifically, during the fine-tuning process, we apply adversarial perturbations to the model weights for regularization, where these perturbations are generated in the opposite direction of gradient descent. Intuitively, WRF4CIR increases the difficulty of fitting the training data, which helps mitigate overfitting in CIR under limited triplet supervision. Extensive experiments on benchmark datasets demonstrate that WRF4CIR significantly narrows the generalization gap and achieves substantial improvements over existing methods.
\end{abstract}




\maketitle

\section{Introduction}
Composed image retrieval (CIR) is a task that retrieves target images based on reference images and modification texts. By integrating visual and textual information, the CIR model enables more accurate and flexible retrieval than traditional retrieval methods, showing great potential in real-world applications like e-commerce \citep{wu2021fashion, goenka2022fashionvlp}, medical diagnostics \citep{cao2014medical}, and document search \citep{hassan2013multi}. The CIR framework typically involves two key components: feature encoder and multimodal fusion module. In general, the reference image and text are first processed by feature encoders, then their features are fused into a unified representation to retrieve the target image.

The feature encoder in previous CIR methods \citep{vo2019composing,chawla2021leveraging,delmas2022artemis} primarily relied on traditional models, such as ResNet \citep{he2016deep} for image and LSTM \citep{hochreiter1997long} for text. Due to the limited multimodal feature encoding capabilities of traditional models, these methods can only achieve suboptimal performance. To overcome this limitation, recent research \citep{baldrati2023composed,bai2023sentence, tian2025ccin, li2025learning,wangtowards} employ vision-language pre-trained models (VLPs) as feature encoders. They leverage the powerful feature extraction and cross-modal alignment capabilities of VLPs, yielding significant improvements in CIR benchmark. However, such gains are often undermined by overfitting during fine-tuning, which has been largely overlooked in previous CIR research.

In this work, we investigate the overfitting issue in VLP-based CIR task. First, we argue that overfitting is a prevalent phenomenon in VLP-based CIR models. To substantiate this, we fine-tune different pre-trained models and visualize the learning curves across a range of CIR methods and datasets. We observe
a progressively widening gap in recall between the training and test sets, which eventually manifests as a significant generalization gap under all experimental settings. These results suggest that overfitting is a general and widespread issue in VLP-based CIR task. To further analyze the overfitting phenomenon, we partition the CIR datasets into different proportions and fine-tune pre-trained models on subsets of varying sizes. We observe a clear correlation between the size of the training data and the generalization gap: the smaller the training set, the larger the gap. The experimental results indicate that the limited amount of training data in CIR contributes to the overfitting observed during fine-tuning of pre-trained models. We also find that commonly used data augmentation methods fail to adequately address this issue, highlighting the challenge of fine-tuning VLP models on limited CIR triplets. At the same time, annotating data for CIR task is both time-consuming and labor-intensive, which raises a key question: 

\begin{center}
\textit{
how can we effectively fine-tune pre-trained models with limited CIR training data?}
\end{center}

\begin{figure*}[t]
    \centering
    \begin{minipage}[t]{0.48\textwidth}
        \centering
        \includegraphics[width=\textwidth]{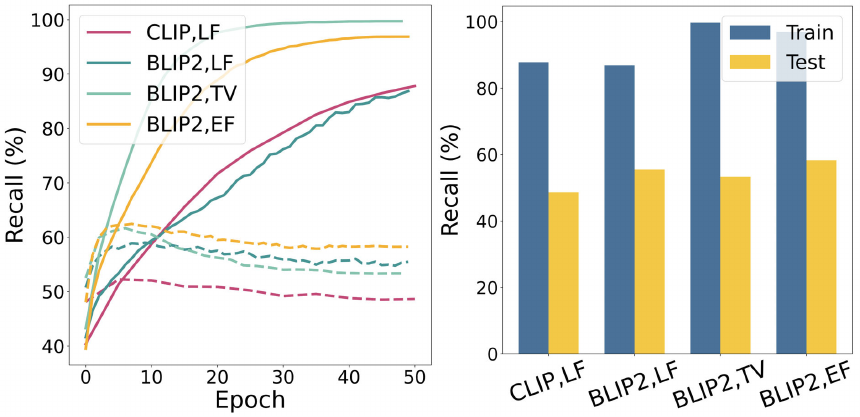}
        \caption*{(a) Across different baseline methods (Solid line: training set; Dashed line: val set)}
    \end{minipage}
    \hfill 
    \begin{minipage}[t]{0.48\textwidth}
        \centering
        \includegraphics[width=\textwidth]{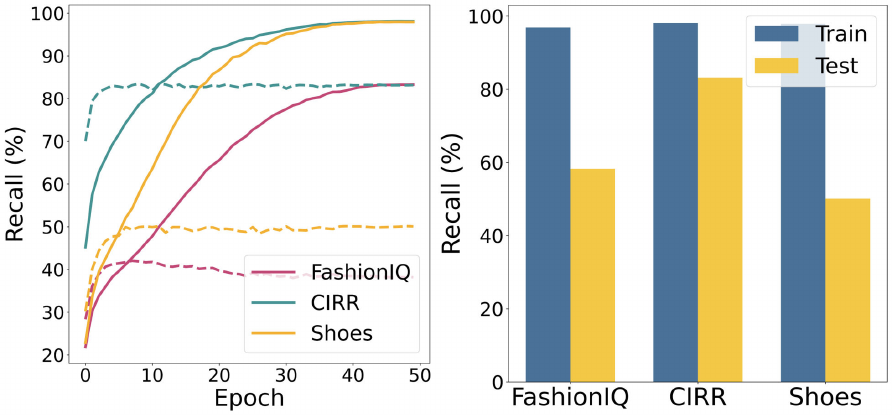}
        \caption*{(b) Across different datasets (Solid line: training set; Dashed line: val set)}
    \end{minipage}
    \caption{Visualization of learning curve and generalization gap across (a) different baseline methods on FashionIQ (LF: late-fusion; EF: early-fusion; TV: sentence-level text inversion); and (b) different datasets, including FashionIQ, CIRR and Shoes. These results suggest that overfitting is a prevalent phenomenon in VLP-based CIR models.}
    \Description{Visualization of learning curve and generalization gap across (a) different baseline methods on FashionIQ (LF: late-fusion; EF: early-fusion; TV: sentence-level text inversion); and (b) different datasets, including FashionIQ, CIRR and Shoes. Most methods achieve over 90\% average recall on the FashionIQ training set upon convergence, while their average recall on the validation set plateaus at around 50\%. Similar phenomena are observed across different datasets, as shown in (b). These results suggest that overfitting is a prevalent phenomenon in VLP-based CIR models.}
    \label{fig1}
\end{figure*}
To address this issue, motivated by prior work \citep{wu2020adversarial,foret2020sharpness} that mitigates overfitting by flattening the weight loss landscape, we introduce a Weight-Regularized Fine-tuning network for Composed Image Retrieval, termed WRF4CIR. It regularizes the fine-tuning process by introducing adversarial perturbations to the model weights. Specifically, in a minibatch, we apply an adversarial perturbation to the model weights in the direction opposite to gradient descent, and then remove it after the parameter update to maintain training stability. Intuitively, these adversarial perturbations increase the model’s learning difficulty on the training data, thus can effectively mitigate overfitting. Experimental results indicate that WRF4CIR utilizes CIR triplets more effectively, achieving full-data performance on FashionIQ with no more than 60\% of the training set. The effectiveness of the proposed method is further validated across different pre-trained models, fusion strategies, fine-tuning strategies, and benchmark datasets. Extensive experiments demonstrate that our method achieves substantial improvements over existing methods without relying on large language models or additional training data. In summary, our main contributions are as follows:
\begin{itemize}[leftmargin=*, itemsep=0pt]
\item We present a systematic study of overfitting in VLP-based CIR, revealing a significant and previously overlooked generalization gap across different models and datasets. 
\item We introduce a weight-regularized fine-tuning network for composed image retrieval, which mitigates overfitting under limited CIR triplet data.
\item Extensive experiments demonstrate that our method achieves substantial improvements over existing approaches across multiple benchmarks and experimental settings.
\end{itemize}

\section{Related Work}
\label{sec:relatedwork}
\subsection{Composed Image Retrieval}
Composed image retrieval (CIR) \citep{vo2019composing, liu2021image, baldrati2023composed, zhang2025composed, song2025comprehensive, psomas2025instance} aims to retrieve a target image by jointly leveraging a reference image and a textual modification. Previous approaches \citep{chawla2021leveraging,kim2021dual,lee2021cosmo,yang2021cross,delmas2022artemis} mainly utilized conventional networks (e.g., ResNet \citep{he2016deep} or LSTM \citep{hochreiter1997long}) as feature encoders, but their performance is suboptimal due to limited multimodal feature encoding capability. Recent methods \citep{wen2023target,bai2023sentence,chen2024fashionern,jiang2024cala, chen2025mai, chen2025offset, shi2025multi} have overcome this limitation by leveraging the powerful representation capabilities of vision-language pre-training models (VLPs), achieving significantly improved results. For example, CLIP4CIR \citep{baldrati2023composed} is the first to employ CLIP \citep{radford2021learning} for extracting image and text features, which are then integrated via a gating mechanism. \citet{bai2023sentence} use the Q-former from BLIP-2 \citep{li2023blip} and enable fusion at the input level by projecting images into the word embedding space of the VLPs. Despite the significant performance gains achieved by VLP-based CIR methods, we find that existing methods commonly suffer from severe overfitting during the fine-tuning of pre-trained models. Several existing methods \citep{feng2024improving,ge2025llm, jang2025visual} have leveraged multi-modal large language models (MLLMs) to construct triplets for CIR. For example, SPN4CIR \citep{feng2024improving} uses MLLMs to generate additional triplets for positive sample enhancement. In contrast to these approaches that rely on extra data, our work focuses on the fine-tuning strategy for pre-trained models on relatively limited CIR data, aiming to mitigate overfitting through weight regularization and improve retrieval performance.

\subsection{Vision-Language Pre-Training Models}
Vision-language pre-training models trained on large-scale datasets can effectively align visual and textual information. Various model architectures \citep{jia2021scaling,wang2021simvlm,wang2023image} and pre-training objectives \citep{radford2021learning,li2021align,li2023blip} have been introduced over time, continuously enhancing multimodal encoding capabilities and generalization on diverse downstream tasks \citep{mokady2021clipcap,song2022clip,sun2024clip}.  Recently, CLIP \citep{radford2021learning}, BLIP \citep{li2022blip}, and BLIP-2 \citep{li2023blip} have been widely adopted in composed image retrieval methods. Their strong feature extraction and cross-modal alignment capabilities have led to substantial improvements on CIR benchmarks. In this work, we adopt these pre-trained models as backbones to systematically investigate the overfitting issue in vision-language pretraining VLP-based CIR tasks. Furthermore, we introduce a weight-regularized fine-tuning network that effectively mitigates this problem across different pre-trained models and datasets.
\subsection{Overfitting and Generalization}
Deep models have demonstrated remarkable capability in learning rich representations for complex tasks; however, they are often prone to overfitting across a variety of scenarios. In image classification, overfitting has been widely explored across different settings, such as standard training regimes \citep{zhang2016understanding,dinh2017sharp,kawaguchi2017generalization}, adversarial training \citep{rice2020overfitting,kim2021understanding}, and few-shot scenarios \citep{sun2019meta,tao2020few}. Beyond classification, overfitting has also been investigated in tasks such as object detection \citep{cai2018cascade}, image segmentation \citep{li2020analyzing}, and cross-modal retrieval \citep{wang2017adversarial}. Despite these efforts, the overfitting phenomenon has been largely overlooked in VLP-based CIR, a multimodal retrieval task involving the fine-tuning of large vision-language models on triplet data. To the best of our knowledge, this work is the first to systematically investigate overfitting in VLP-based CIR, revealing that it is prevalent across existing methods and represents a key bottleneck limiting retrieval performance.
\begin{table}[t]
\centering
\caption{Average Recall of the baseline SPRC under various data augmentation methods. Most of them show limited effectiveness except for SPN, which constructs additional triplets for CIR. (CO: CutOut; TA: TrivialAugment; AM: AugMix; SPN: SPN4CIR.)}
\label{tabDA}
\resizebox{0.9\columnwidth}{!}{\begin{tabular}{lccccc} 
\toprule
Dataset & SPRC & +CO & +TA & +AM & +SPN  \\
\midrule
CIRR & 83.02 & 82.69 & 82.65 & 82.70 &  83.56\\
FashionIQ & 65.15 & 65.36 & 65.46 & 65.41 &  66.41\\
\bottomrule
\end{tabular}}
\end{table}

\begin{figure}[!t]
\centering
\includegraphics[width=0.82\columnwidth]{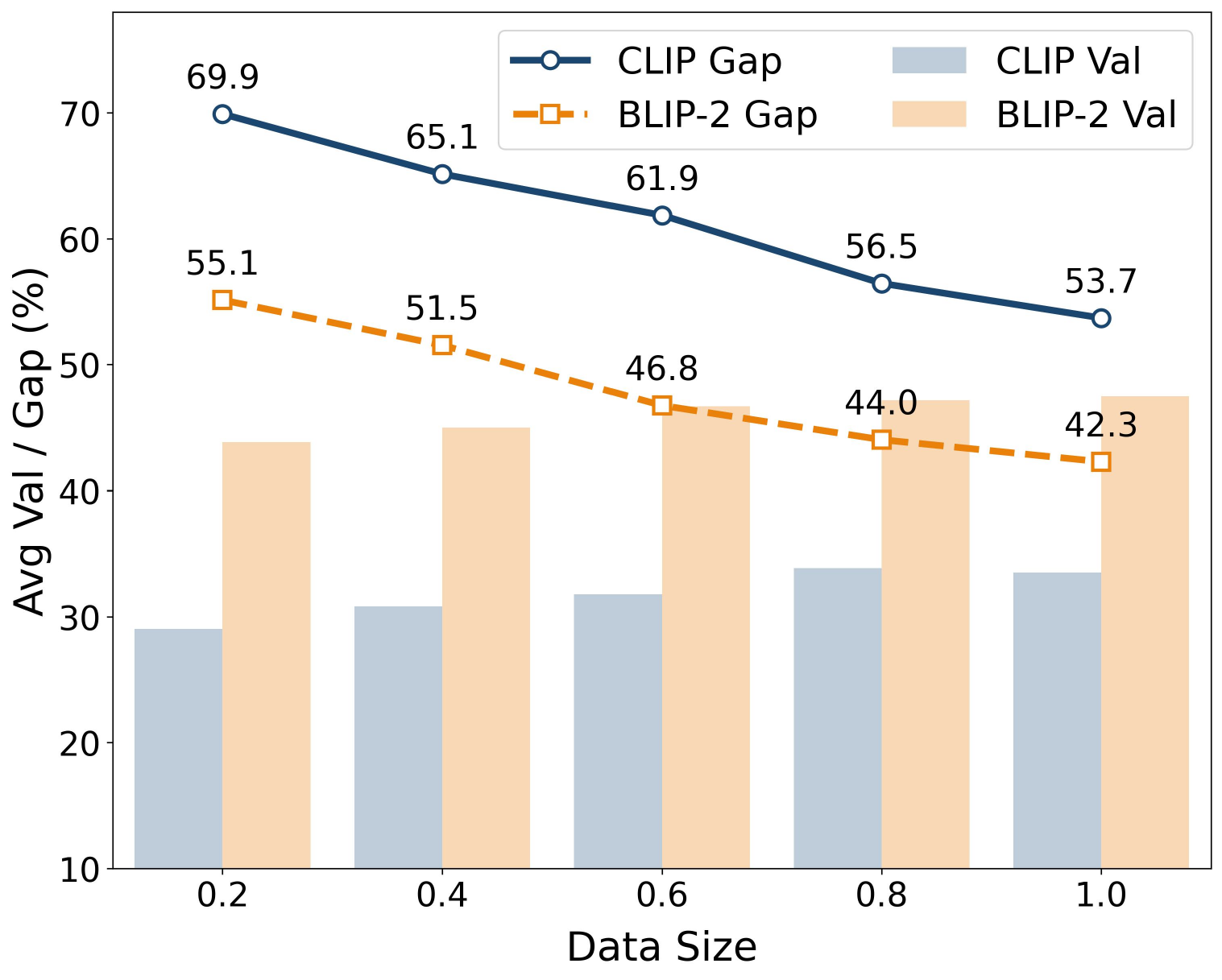}
\caption{Generalization gap and recall on FashionIQ under varying training data sizes. Limited training data contributes to overfitting during fine-tuning.}
\Description{Generalization gap and recall on FashionIQ under varying training data sizes. It is observed that as the size of the training data decreases, the validation recall drops and the generalization gap continues to widen. These experimental results indicate that the limited amount of training data contributes to the overfitting phenomenon during the fine-tuning of pre-trained models.}
\label{fig2}
\end{figure}

\section{Overfitting Issue in VLP-Based Composed Image Retrieval}
\label{sec:method1}
In this section, we first visualize the overfitting phenomenon in VLP-based composed image retrieval. We then provide a detailed analysis of this overfitting issue, and finally introduce the motivation behind our proposed method.

\subsection{Visualization} 
In VLP-based CIR, the fine-tuning process is typically monitored using validation recall, potentially overlooking the model’s behavior on the training set. To more comprehensively characterize the fine-tuning process, we record and report the model’s recall on both the training and validation sets, which better reflects the model’s generalization ability. Specifically, we conduct experiments using widely adopted VLP backbones and fusion strategies from recent state-of-the-art baselines \citep{baldrati2023composed,jiang2024cala,tian2025ccin}. We visualize the learning curves and generalization gaps across different baseline methods and datasets (FashionIQ \citep{wu2021fashion}, CIRR \citep{liu2021image}, and Shoes \citep{guo2018dialog}), and the results are presented in \Cref{fig1}. As illustrated in \Cref{fig1}(a), most methods achieve over 90\% average recall on the FashionIQ training set upon convergence, while their average recall on the validation set plateaus at around 50\%. Notably, the best validation performance typically occurs within the first five epochs—well before the model fully converges on the training set. Further training continues to improve training recall but yields no gain on the validation set, indicating clear signs of overfitting. Similar phenomena are observed across different datasets, as shown in \Cref{fig1}(b). Overall, we consistently observe a significant generalization gap under all experimental settings, indicating that overfitting is a prevalent phenomenon in VLP-based CIR task.

\subsection{Analysis about the overfitting issue} To further analyze this overfitting phenomenon, we fine-tune pre-trained models with varying sizes of training data. Specifically, we randomly split the training sets into subsets of different sizes with proportions \{0.2, 0.4, 0.6, 0.8, 1.0\} and conduct experiments under three datasets (FashionIQ, CIRR, and Shoes) and two pre-trained models (CLIP and BLIP-2). The generalization gap and validation recall on the FashionIQ dataset is shown in \Cref{fig2}, and detailed results for each dataset and additional model are provided in the supplementary material. It is observed that as the size of the training data decreases, the validation recall drops and the generalization gap continues to widen. These experimental results indicate that the limited amount of training data contributes to the overfitting phenomenon during the fine-tuning of pre-trained models. Furthermore, we evaluated three widely adopted data augmentation methods, including CutOut \citep{devries2017improved}, AugMix \citep{hendrycks2019augmix}, and TrivialAugment \citep{muller2021trivialaugment}, alongside the recent SPN4CIR \citep{feng2024improving}. As shown in \Cref{tabDA}, most of these methods demonstrate limited effectiveness on top of the competitive SPRC baseline \citep{bai2023sentence}, with the exception of SPN4CIR, which constructs additional triplets for CIR. However, CIR data generated by large language models (LLMs) often suffer from inconsistent quality \citep{feng2024improving,levy2024data,zhou2025scale}. Meanwhile, annotating data for the CIR task remains both time-consuming and labor-intensive. These factors highlight the challenge of overfitting in VLP-based CIR and naturally raises the key question of how to effectively fine-tune pre-trained models with limited CIR training data.
\begin{figure*}[tb] \centering
    \includegraphics[width=0.8\textwidth]{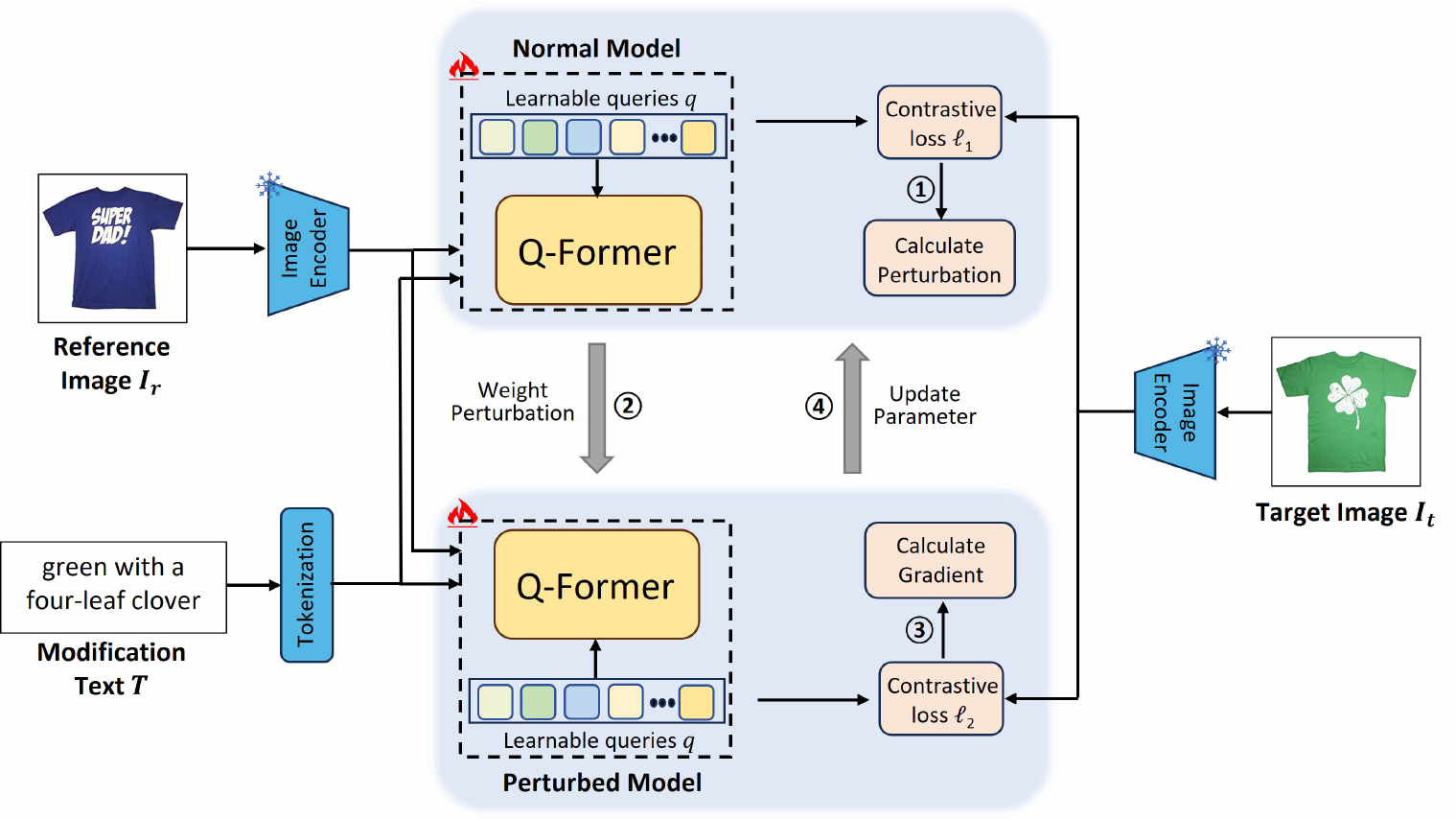}
    \caption{\textbf{Illustration of our WRF4CIR framework.} The reference image and modification text are first fused through the Q-Former, then aligned with the target image by optimizing a contrastive loss. Building on this process, adversarial perturbations are generated and applied to the model weights for regularization. Finally, the gradients computed from the perturbed model are used to update the parameters of the normal model.}
    \Description{Illustration of our WRF4CIR framework. The reference image and modification text are first fused through the Q-Former, then aligned with the target image by optimizing a contrastive loss. Building on this process, adversarial perturbations are generated and applied to the model weights for regularization. Finally, the gradients computed from the perturbed model are used to update the parameters of the normal model.}
    \label{fig3}
\end{figure*}

\subsection{Motivation of WRF4CIR} 
To address this issue, a common approach is to employ regularization techniques, since they can effectively reduce the risk of overfitting caused by limited training data and excessive model complexity \citep{geman1992neural,belkin2019reconciling}. Furthermore, inspired by previous studies \citep{wu2020adversarial,foret2020sharpness} that improve generalization by flattening the weight loss landscape, we introduce a weight-regularized fine-tuning network for CIR, named WRF4CIR, which regularizes the fine-tuning process by applying perturbations to the model weights. The optimization objective of WRF4CIR is formulated as follows:\begin{equation}
\label{equ1}
    \min _{\mathbf{\theta}} \{\mathcal{L}(\mathbf{\theta})+(\mathcal{L}(\mathbf{\theta}+\mathbf{\delta})-\mathcal{L}(\mathbf{\theta}))\} \rightarrow \min _{\mathbf{\theta}} \mathcal{L}(\mathbf{\theta}+\mathbf{\delta}).
\end{equation}
Here, $\mathcal{L}(\mathbf{\theta})$ denotes the standard optimization objective of the model, which, in the context of CIR, corresponds to the contrastive loss between the target representation and the query representation. The term $(\mathcal{L}(\mathbf{\theta}+\mathbf{\delta}) - \mathcal{L}(\mathbf{\theta}))$ serves as a regularization component that reflects the flatness of the weight loss landscape, where $\mathbf{\delta}$ represents a perturbation applied to the model weights. In the next section, we introduce the implementation details of WRF4CIR.

\section{Weight-Regularized Fine-Tuning Network for Composed Image Retrieval}
\label{sec4}
\subsection{Basic framework} Composed image retrieval (CIR) aims to retrieve target images $I_t$ from a large-scale database using multimodal queries \{$I_r, T$\}. $I_r$ and $T$ are processed by given feature encoders, after which their features are fused into a unified representation to retrieve the target image $I_t$. In this part, we first present a basic CIR framework based on BLIP-2, which consists of an image encoder and a lightweight Q-Former, as shown in \Cref{fig3}. Specifically, the framework employs a frozen image encoder to extract initial features from both the reference image $I_r$ and the target image $I_t$. These features are then passed to the Q-Former, where they interact with learnable query embeddings $q$ and tokenized text, producing fused query representations $u$ and target image representations $v$. Finally, the fused query representation $u$ and the target representation $v$ are aligned by optimizing a contrastive loss, which aims to minimize the distance between positive pairs while maximizing it for negative pairs. The contrastive objective is formulated as:
\begin{equation}
\label{equ2}
     \ell_{q 2 t}\left(\hat{u}, \hat{v}\right)=-\frac{1}{|\mathcal{B}|} \sum_{i \in \mathcal{B}} \log \frac{\exp \left(\tau \hat{u}_i^{T} \hat{v}_i\right)}{\sum_{j \in \mathcal{B}} \exp \left(\tau \hat{u}_i^{T} \hat{v}_j\right)}.
\end{equation}
Here, ${\hat{u}}$ and ${\hat{v}}$ are the normalized vectors of $u$ and $v$, $\mathcal{B}$ is a randomly sampled mini-batch, and $\tau$ is a temperature coefficient for scaling in contrastive loss. Notably, our method is applicable to different pre-trained models, fusion strategies, and fine-tuning strategies, which we discuss in the experimental section. \begin{table*}[t]
\centering
\captionsetup{position=top}
\caption{\textbf{Results on FashionIQ val set.} The best and second-best results are marked in \textcolor{red}{Red} and \textcolor{blue}{Blue}, respectively. The symbol \textsuperscript{\textdagger} indicates methods that leverage additional data.}
\resizebox{0.8\textwidth}{!}{
   \begin{tabular}{ll cc|cc|cc|ccc} 
        \toprule
        \multirow{2}{*}{Method} & \multirow{2}{*}{Backbone} & \multicolumn{2}{c}{$\textbf{Dress}$} & \multicolumn{2}{c}{$\textbf{Shirt}$} & \multicolumn{2}{c}{$\textbf{TopTee}$} & \multicolumn{3}{c}{$\textbf{Average}$} \\
        \cmidrule(r){3-4} \cmidrule(lr){5-6} \cmidrule(lr){7-8} \cmidrule(l){9-11}
        & & R10 & R50 & R10 & R50 & R10 & R50 & R10 & R50 & Rmean \\
        \midrule
        CLIP4CIR \citep{baldrati2023composed} & CLIP ResNet-50 & 37.67 & 63.16 & 39.87 & 60.84 & 44.88 & 68.59 & 40.80 & 64.20 & 52.50 \\
        BLIP4CIR+Bi \citep{liu2024bi} & BLIP ViT-B & 42.09 & 67.33 & 41.76 & 64.28 & 46.61 & 70.32 & 43.49 & 67.31 & 55.40 \\
        CASE \citep{levy2024data} & BLIP ViT-B & 47.44 & 69.36 & 48.48 & 70.23 & 50.18 & 72.24 & 48.79 & 70.68 & 59.74 \\
        CaLa \citep{jiang2024cala} & BLIP-2 ViT-L & 42.38 & 66.08 & 46.76 & 68.16 & 50.93 & 73.42 & 46.69 & 69.22 & 57.95 \\
        Re-ranking \citep{liu2023candidate} & BLIP ViT-B & 48.14 & 71.43 & 50.15 & 71.25 & 55.23 & 76.80 & 51.17 & 73.13 & 62.15 \\
        SPRC \citep{bai2023sentence} & BLIP-2 ViT-G & 49.18 & 72.43 & 55.64 & 73.89 & 59.35 & 78.58 & 54.92 & 74.97 & 64.85 \\
        CIR-LVLM \citep{sun2024leveraging} & Qwen-7B & 50.42 & 73.57 & \textcolor{blue}{\textbf{58.59}} & \textcolor{blue}{\textbf{75.86}} & 59.61 & 78.99 & 56.21 & 76.14 & 66.17 \\
        CCIN \citep{tian2025ccin} & BLIP-2 ViT-G & 49.38 & 72.58 & 55.93 & 74.14 & 57.93 & 77.56 & 54.41 & 74.76 & 64.59 \\
        TME \citep{li2025learning} & BLIP-2 ViT-G & 49.73 & 71.69 & 56.43 & 74.44 & 59.31 & 78.94 & 55.15 & 75.02 & 65.09\\
        RUNC \citep{wangtowards} & BLIP-2 ViT-G & 48.93 & 73.53 & 57.26 & 75.32 & 60.38 & 79.86 & 55.52 & 76.23 & 65.88\\
        \textbf{WRF4CIR (ours)} & BLIP-2 ViT-G & \textcolor{blue}{\textbf{51.71}} & \textcolor{red}{\textbf{75.41}} & \textcolor{red}{\textbf{58.94}} & \textcolor{red}{\textbf{76.21}} & \textcolor{red}{\textbf{61.40}} & \textcolor{red}{\textbf{80.37}} & \textcolor{red}{\textbf{57.35}} & \textcolor{red}{\textbf{77.33}} & \textcolor{red}{\textbf{67.34}} \\
        \cmidrule{1-11}
        DQU-CIR\textsuperscript{\textdagger} \citep{wen2024simple} & CLIP ViT-H & \textcolor{red}{\textbf{51.9}} & \textcolor{blue}{\textbf{74.37}} & 53.57 & 73.21 & 58.48 & 79.23 & 54.65 & 75.60 & 65.12 \\
        SPRC+SPN\textsuperscript{\textdagger} \citep{feng2024improving} & BLIP-2 ViT-G & 50.57 & 74.12 & 57.70 & 75.27 & \textcolor{blue}{\textbf{60.84}} & \textcolor{blue}{\textbf{79.96}} & \textcolor{blue}{\textbf{56.37}} & \textcolor{blue}{\textbf{76.45}} & \textcolor{blue}{\textbf{66.41}} \\
        \bottomrule
   \end{tabular}
}
\label{table FashionIQ}
\end{table*}

\subsection{Weight perturbation strategy} 
The proposed WRF4CIR introduces perturbations to the model weights in order to regularize the weight loss landscape and mitigate overfitting. The optimization objective of WRF4CIR has been defined in Equation (\ref{equ1}). Here, we mainly discuss the choice of perturbation strategy in WRF4CIR.
Although random weight perturbation offers a simple implementation, its arbitrary direction may interfere with model optimization and lead to suboptimal performance. To overcome this limitation, we adopt a gradient-based adversarial perturbation strategy, similar to AWP \citep{wu2020adversarial}, where perturbations are applied to the model weights in the direction opposite to their gradients. Intuitively, these adversarial perturbations increase the learning difficulty on the training data, which facilitates fine-tuning on limited CIR triplets and helps mitigate overfitting. Thus, the optimization objective of WRF4CIR can be further expressed as follows:
\begin{equation}
\label{equ3}
\begin{split}
    &\min_{\theta} \max_{\delta} \mathcal{L}(\theta+\delta) \rightarrow \\
    &\min_{\theta} \max_{\delta}  \ell_{q 2 t}\left(f_{\theta+\delta}\left(I_r, T\right), f_{\theta+\delta}\left(I_t\right)\right) \\
    &\text{s.t.} \quad \|\delta_l\| \leq \gamma \|\theta_l\|.
\end{split}
\end{equation}
Here, $\theta$ denotes the model weights, $\delta$ denotes the weight perturbation, and $\theta_l$ ($\delta_l$) denotes the model weights (weight perturbations) in the $l$-th layer. The perturbed model is denoted by $f_{\theta+\delta}$. For the perturbation constraint, we follow the setup adopted in the adversarial training scenario \citep{wu2020adversarial}, where the magnitude of the weight perturbation is kept proportional to the norm of the model weights in each network layer. This helps to determine a better weight perturbation for different layers of the network. Meanwhile, a hyperparameter $\gamma$ is introduced to flexibly control the overall perturbation strength.

\subsection{Weight-regularized fine-tuning} Building upon the basic CIR framework and the weight perturbation strategy, the overall pipeline of WRF4CIR is illustrated in \Cref{fig3}.
Specifically, WRF4CIR performs two forward-backward propagation per mini-batch during training: one for the normal model and one for the perturbed model. For the normal model, we follow the process illustrated in the basic framework and compute the model gradients to determine the direction of weight perturbations. Based on the model gradients, adversarial perturbations $\delta$ are generated and constrained according to the norm of model weights and the predefined perturbation strength $\gamma$, as shown in the following equation:
\begin{equation}
\label{equ4}
    \delta \leftarrow \gamma \frac{\nabla_{\theta} \ell_{q 2 t}\left(f_{\theta}\left(I_r, T\right), f_{\theta}\left(I_t\right)\right)}{\left\|\nabla_{\theta} \ell_{q 2 t}\left(f_{\theta}\left(I_r, T\right), f_{\theta}\left(I_t\right)\right)\right\|}\|\theta\|.
\end{equation}
The weight perturbation $\delta$ is then added to the normal model to obtain the perturbed model, denoted as $f_{\theta + \delta}$. Finally, we optimize the perturbed model and use its gradients to update the parameters of the normal model, which is equivalent to removing the weight perturbations after optimizing the perturbed model, as illustrated in the following equation:
\begin{equation}
\label{equ5}
    \theta \leftarrow(\theta+\delta)-\eta \nabla_{\theta+\delta} \ell_{q 2 t}\left(f_{\theta+\delta}\left(I_r, T\right), f_{\theta+\delta}\left(I_t\right)\right)-\delta.
\end{equation}
The complete procedure of WRF4CIR is summarized in \Cref{alg1}, where the weight regularization module is portable and can be seamlessly integrated into different VLP models and fusion strategies. 
\begin{algorithm}
\caption{WRF4CIR}
\label{alg1}
\begin{algorithmic}[1]
   \State \textbf{Input:} Training set $D_t = \{ (I_r^i, T^i, I_t^i) \}_{i=1}^n$, perturbation strength $\gamma$, learning rate $\eta$, normal model $f$.
   \State \textbf{Output:} Fine-tuned model $f$.
   \For{$t = 0$ \textbf{to} $T-1$}
      \State Forward propagation and calculate gradients of normal \State model:
      \State \quad $L(f) \gets \nabla_{f_\theta}\ell_{q2t}(f_\theta(t;T,I_r), f_\theta(t;I_t))$
      \State Add adversarial perturbation:
      \State \quad $f_{\theta+\delta}(t) \gets f_\theta(t) + \gamma\frac{L(f)}{\|L(f)\|}\|f_\theta(t)\|$
      \State Forward propagation and calculate gradients of perturbed 
      \State model:
      \State \quad $L'(f) \gets \nabla_{f_{\theta+\delta}}\ell_{q2t}(f_{\theta+\delta}(t;T,I_r), f_{\theta+\delta}(t;I_t))$
      \State Update parameter of normal model:
      \State \quad $f_\theta(t+1) \gets f_\theta(t)-\eta L'(f)$
   \EndFor
\end{algorithmic}
\end{algorithm}\begin{table*}
\centering
\captionsetup{position=top}
\caption{\textbf{Results on CIRR test set.} The best and second-best results are marked in \textcolor{red}{Red} and \textcolor{blue}{Blue}, respectively. The symbol \textsuperscript{\textdagger} indicates methods that leverage additional data.}
\resizebox{0.75\textwidth}{!}{    
    \begin{tabular}{llcccc|ccc|c} 
        \toprule
        \multirow{2}{*}{Method} & \multirow{2}{*}{Backbone} & \multicolumn{4}{c}{\textbf{Recall@K}} & \multicolumn{3}{c}{\textbf{Recall$_{\text{subset}}$@K}} & \multirow{2}{*}{Avg.} \\
        \cmidrule(r){3-6} \cmidrule(l){7-9}
        & & K=1 & K=5 & K=10 & K=50 & K=1 & K=2 & K=3 & \\
        \cmidrule{1-10}

        CLIP4CIR \citep{baldrati2023composed} & CLIP ResNet-50 & 40.91 & 74.53 & 84.77 & 97.35 & 70.22 & 87.80 & 94.46 & 72.38 \\
        BLIP4CIR+Bi \citep{liu2024bi} & BLIP ViT-B & 40.15 & 73.08 & 83.88 & 96.27 & 72.10 & 88.27 & 95.93 & 72.59 \\
        CASE \citep{levy2024data} & BLIP ViT-B & 48.00 & 79.11 & 87.25 & 97.57 & 75.88 & 90.58 & 96.00 & 77.50 \\
        CaLa \citep{jiang2024cala} & BLIP-2 ViT-L & 49.11 & 81.21 & 89.59 & 98.00 & 76.27 & 91.04 & 96.46 & 78.74 \\
        TG-CIR \citep{wen2023target} & CLIP ViT-B & 45.25 & 78.29 & 87.16 & 97.30 & 72.84 & 89.25 & 95.13 & 75.57 \\
        Re-ranking \citep{liu2023candidate} & BLIP ViT-B & 50.55 & 81.75 & 89.78 & 97.18 & 80.04 & 91.90 & 96.58 & 80.90 \\
        SPRC \citep{bai2023sentence} & BLIP-2 ViT-G & 51.96 & 82.12 & 89.74 & 97.69 & 80.65 & 92.31 & 96.60 & 81.39 \\
        CIR-LVLM \citep{sun2024leveraging}& Qwen-7B &53.64&83.76&90.60&97.93&79.12&92.33&96.67&81.44\\
        CCIN \citep{tian2025ccin} & BLIP-2 ViT-G& 53.41 & \textcolor{blue}{\textbf{84.05}} & \textcolor{blue}{\textbf{91.17}} & 98.00 & - & - & - & -  \\
        TME \citep{li2025learning} & BLIP-2 ViT-G & 53.42 & 82.99 & 90.24 & 98.15 & \textcolor{blue}{\textbf{81.04}} & 92.58 & \textcolor{blue}{\textbf{96.94}} & 82.01\\
        RUNC \citep{wangtowards} & BLIP-2 ViT-G & 53.81 & 83.47 & 91.11 & 98.22 & 80.87 & 92.36 & \textcolor{blue}{\textbf{96.94}} & 82.17\\
        \textbf{WRF4CIR(ours)} & BLIP-2 ViT-G & \textcolor{red}{\textbf{56.58}} & \textcolor{red}{\textbf{85.45}} & \textcolor{red}{\textbf{92.24}} & \textcolor{red}{\textbf{98.68}} & 80.48 & \textcolor{blue}{\textbf{92.60}} & 96.84 & \textcolor{red}{\textbf{82.96}} \\
        \cmidrule{1-10}
        DQU-CIR\textsuperscript{\textdagger} \citep{wen2024simple} & CLIP ViT-H & 46.22 & 78.17 & 87.64 & 97.81 & 70.92 & 87.69 & 94.68 & 74.55 \\
        SPRC+SPN\textsuperscript{\textdagger} \citep{feng2024improving} & BLIP-2 ViT-G & \textcolor{blue}{\textbf{55.06}} & 83.83 & 90.87 & \textcolor{blue}{\textbf{98.29}} & \textcolor{red}{\textbf{81.54}} & \textcolor{red}{\textbf{92.65}} & \textcolor{red}{\textbf{97.04}} & \textcolor{blue}{\textbf{82.69}} \\
        \bottomrule
   \end{tabular}
   }
\label{table CIRR}
\end{table*}

\section{Experiments}
\label{sec:exp}
\subsection{Experimental Setup}
\subsubsection{Implementation details.} Our method is implemented based on PyTorch and runs on a single NVIDIA RTX A100 GPU with 40GB of memory. 
For a fair comparison, we adopt BLIP-2 ViT-G as the backbone, which has been widely adopted in recent state-of-the-art methods \citep{bai2023sentence,tian2025ccin,li2025learning}.
Following the same setup as BLIP-2 based CIR methods \citep{bai2023sentence}, we initialize the image encoder (kept frozen during training) and treat the Q-Former along with learnable queries as trainable parameters. We use the AdamW \citep{loschilov2019decoupled} optimizer with a weight decay of 0.05. The input images are resized to 224 × 224, with a padding ratio of 1.25 to ensure uniformity. The learning rate is initialized to 1e-5 following a cosine schedule for both the \textbf{CIRR} \citep{liu2021image} and \textbf{FashionIQ} \citep{wu2021fashion} datasets. Before applying weight-regularized fine-tuning, we perform a 3-epoch warm-up phase, considering that the model does not exhibit overfitting during the initial training epochs.
\subsubsection{Datasets and metrics.} We evaluate our method on two CIR benchmarks: \textbf{FashionIQ} \citep{wu2021fashion} contains fashion items of three categories: Dress, Shirt, and TopTee. Every triplet consists of a reference image, two relative captions and a target image. We evaluate our model on the validation set, which consists of 6K triplets. Following previous studies \citep{baldrati2023composed,bai2023sentence,tian2025ccin}, we use the recall at rank \textit{K} (R@\textit{K}) as the evaluation metric and calculate the average to assess the overall performance. \textbf{CIRR} \citep{liu2021image} comprises almost 21K real-life open-domain images taken from the {NLVR$^2$} dataset \citep{suhr2018corpus}. We assess our model on the test set of CIRR, which contains 4.1K testing triplets. We use the recall at rank \textit{K} where \textit{K} = 1, 5, 10, 50 and Recallsubset@K as evaluation metrics. The final score is computed as the average of Recall@5 and Recallsubset@1.
\subsection{Performance Comparison}
The experimental results on the FashionIQ dataset, evaluated on the original split of val set, are reported in \Cref{table FashionIQ}. It can be observed that our method achieves the highest recall across most evaluation metrics on FashionIQ, except for Dress R@10, where it ranks second. Compared to VLP-based methods such as SPRC \citep{bai2023sentence} and CCIN \citep{tian2025ccin}, our approach outperforms the best of them (SPRC) by 2.43\% and 2.36\% in average R@10 and R@50, respectively. WRF4CIR even outperforms MLLM-based data augmentation methods (e.g., SPN4CIR \citep{feng2024improving}), achieving a 0.93\% higher average recall without any additional data. These results demonstrate the effectiveness of the proposed method in improving model generalization and enhancing test performance.

We further report results on the CIRR dataset. Note that this dataset is more challenging than FashionIQ, as the degree of overfitting and the generalization gap are much smaller compared to those in FashionIQ. As shown in \Cref{table CIRR}, the proposed approach still outperforms all other methods in terms of average recall. Compared to recent methods that incorporate LLMs, our method can achieve better performance. For example, CIR-LVLM \citep{sun2024leveraging} uses a LLM (Qwen-7B \citep{bai2023qwen}) as a user intent-aware encoder, yet WRF4CIR yields a 1.52\% improvement in average recall even without relying on LLMs. These result further highlights the importance of leveraging the weight-regularized fine-tuning strategy for VLP-based CIR. 
\begin{figure}[t] 
    \centering
    \includegraphics[width=0.45\textwidth]{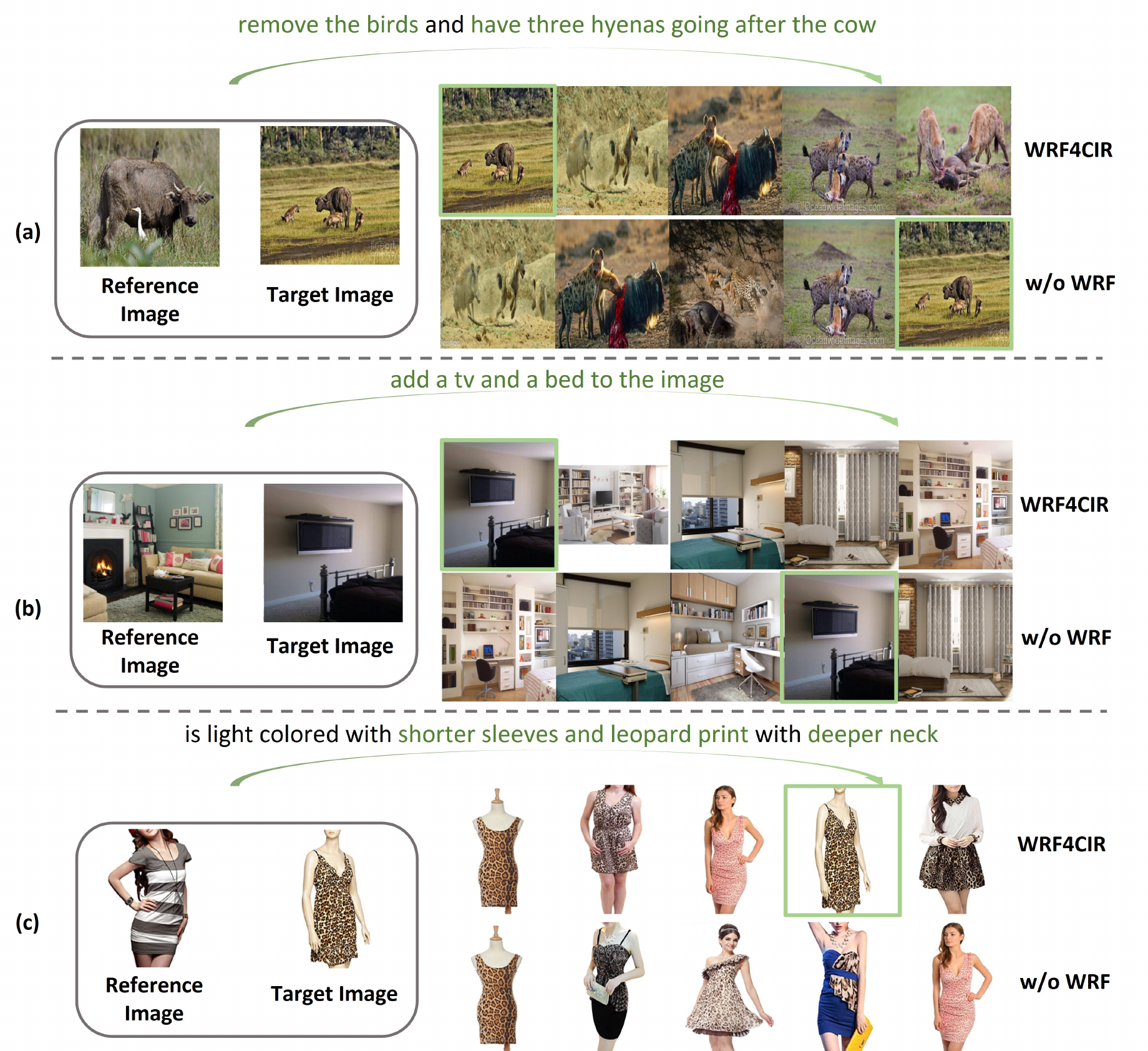}
    \caption{\textbf{Case study between the baseline (w/o WRF) and our WRF4CIR on CIRR ((a) and (b)) and FashionIQ (c). The green boxes indicate the ground-truth targets.} }
    \Description{Case study between the baseline (w/o WRF) and our WRF4CIR on CIRR ((a) and (b)) and FashionIQ (c). The green boxes indicate the ground-truth targets. These results demonstrate that the effectiveness of our method in understanding reference image and text qualitatively, enabling more accurate target image retrieval.}
    \label{fig4}
\end{figure}
\begin{figure*}
    \begin{minipage}[t]{0.248\textwidth}
        \includegraphics[width=\textwidth]{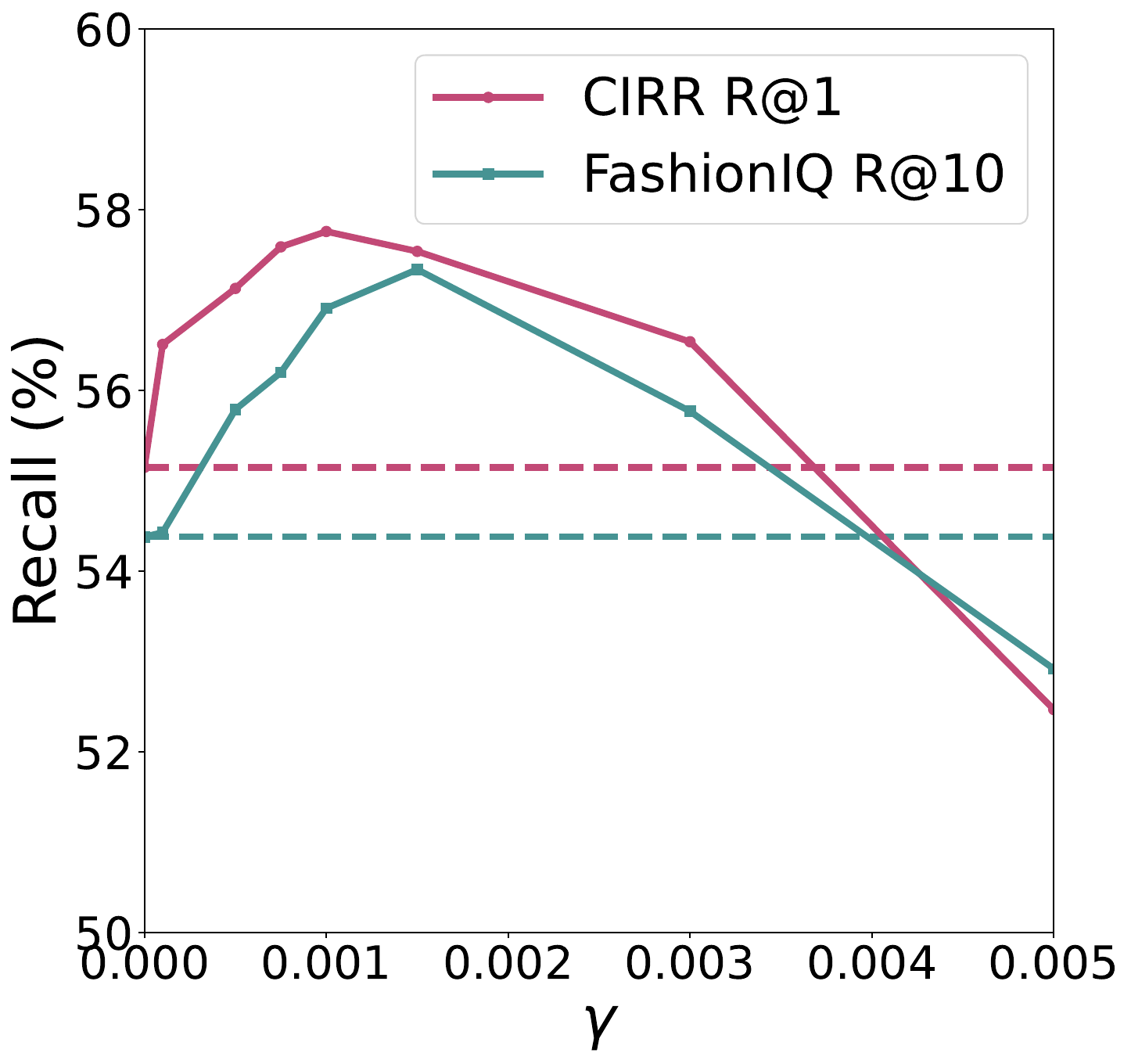}
        \caption*{(a) Weight perturbation}
    \end{minipage}
    \begin{minipage}[t]{0.24\textwidth}
        \includegraphics[width=\textwidth]{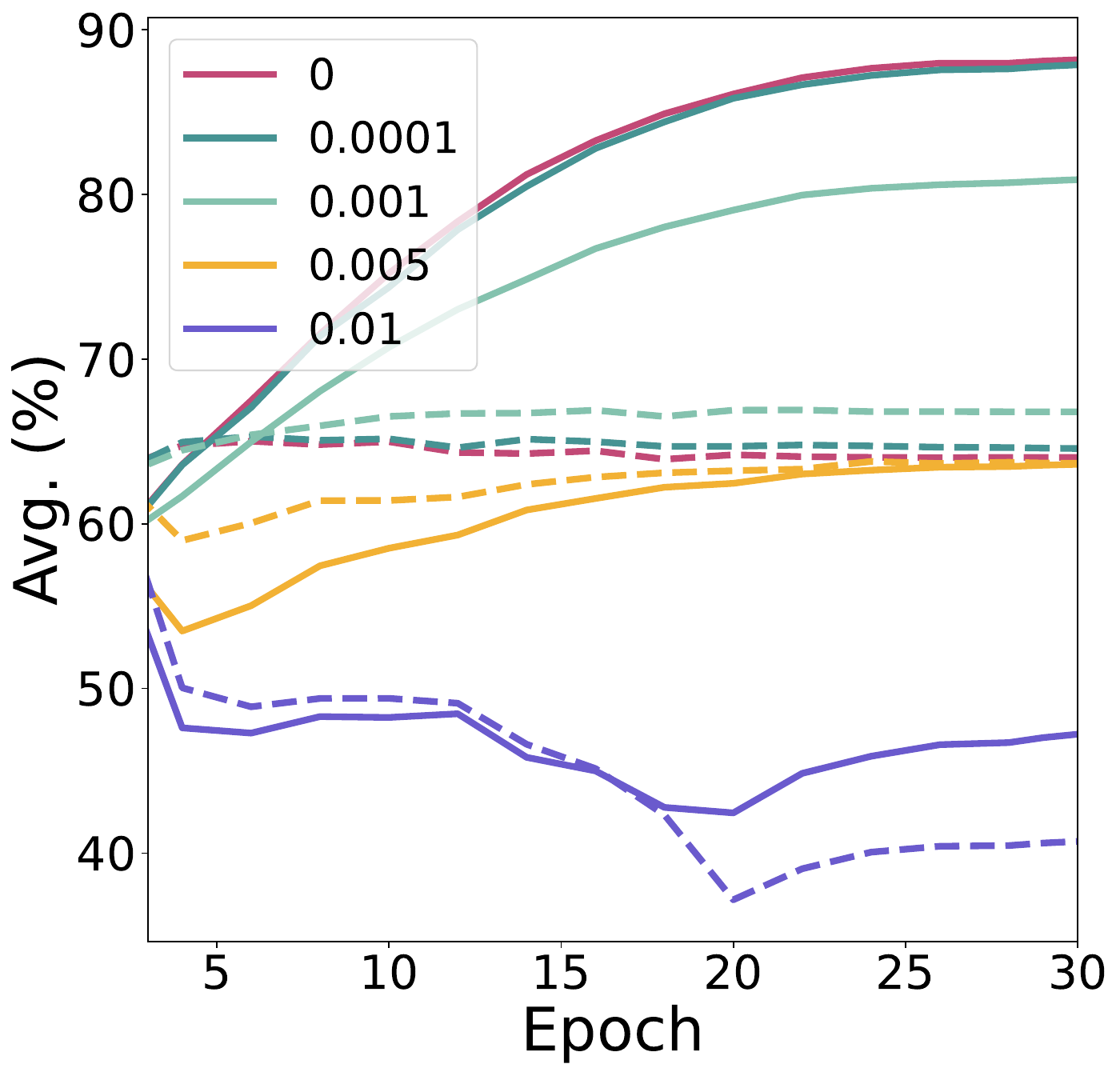}
        \caption*{(b) Learning curve}
    \end{minipage}
    \begin{minipage}[t]{0.253\textwidth}
        \includegraphics[width=\textwidth]{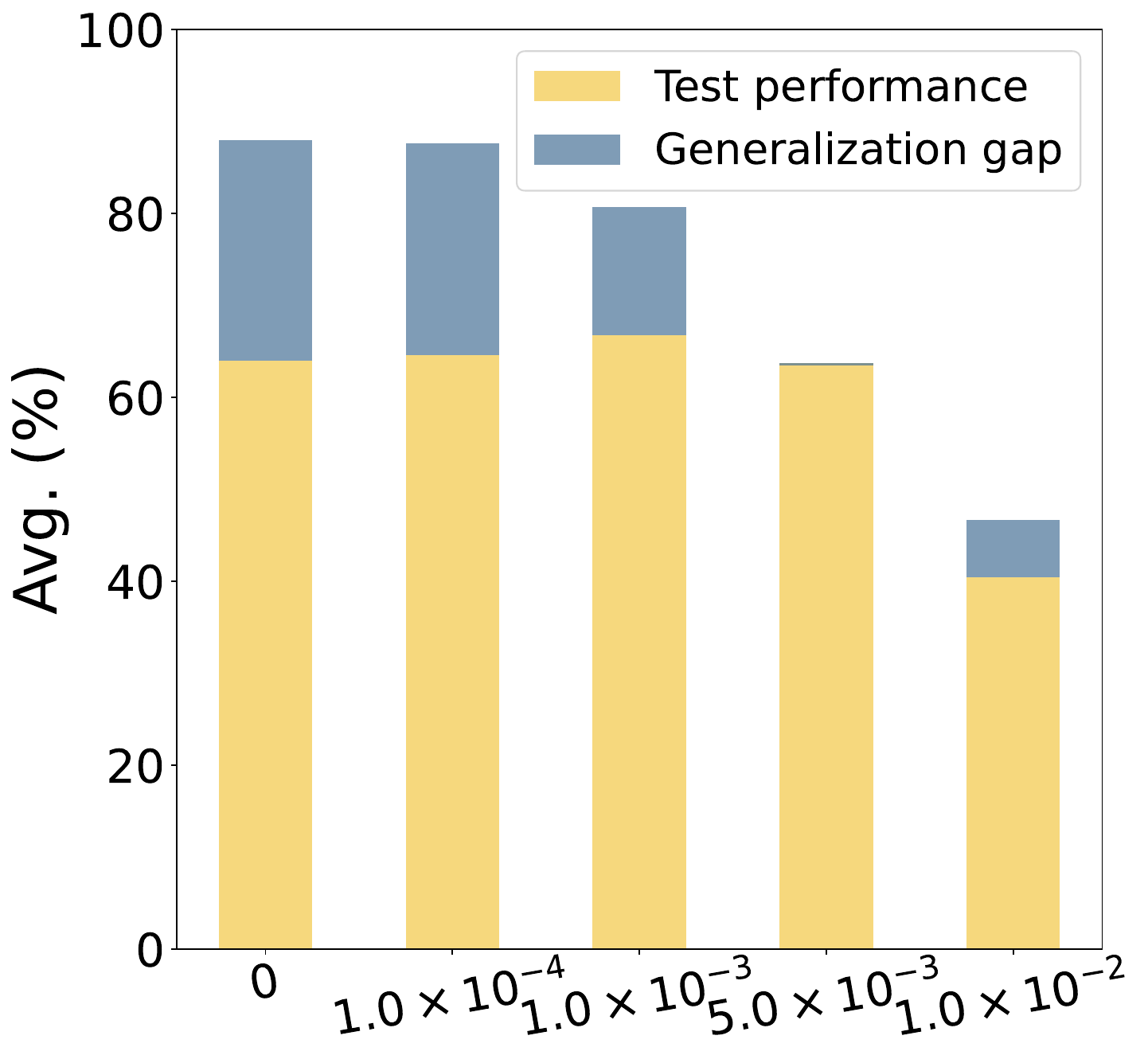}
        \caption*{(c) Generalization gap}
    \end{minipage}
    \begin{minipage}[t]{0.245\textwidth}
        \includegraphics[width=\textwidth]{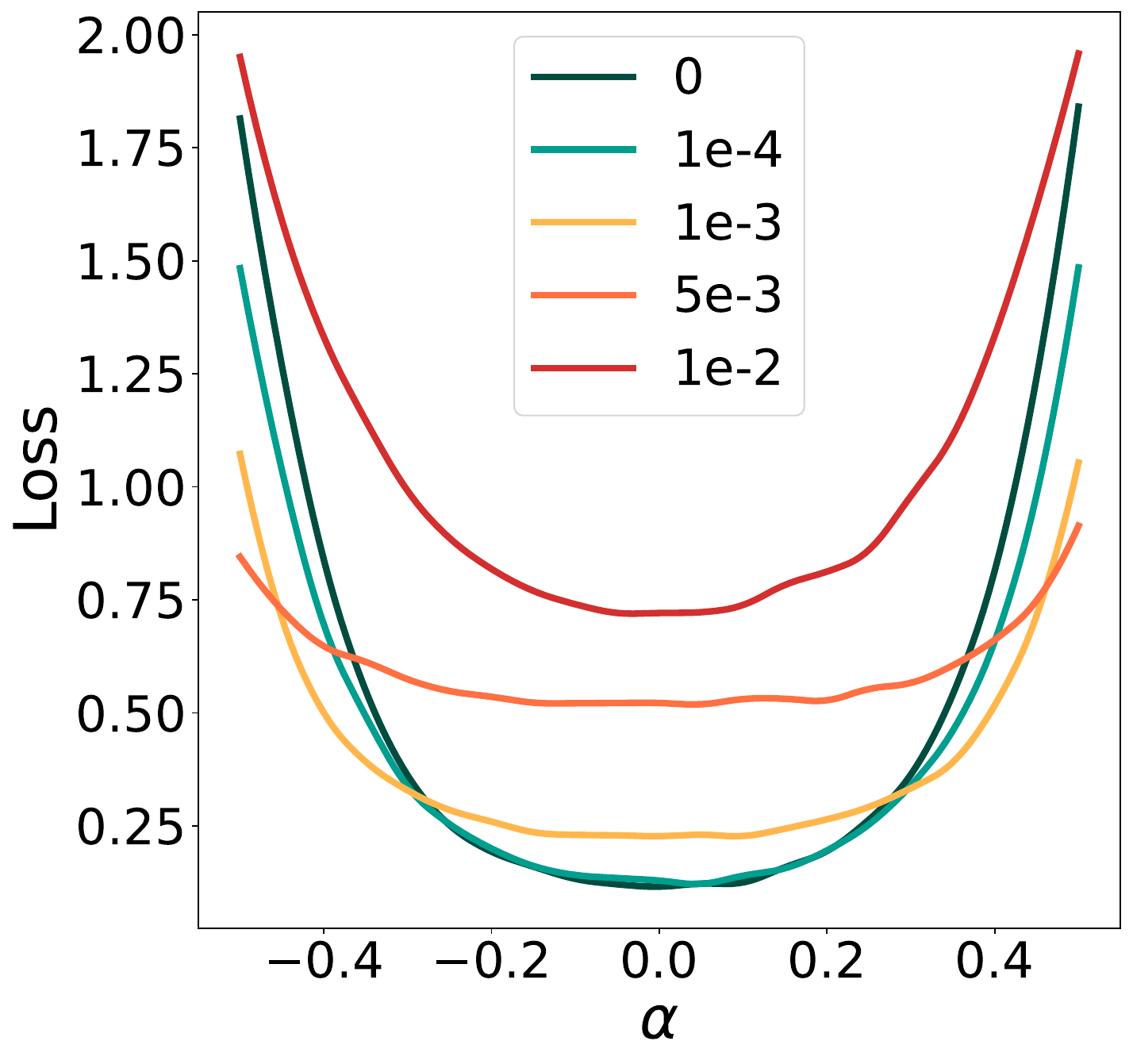}
        \caption*{\shortstack{(d) Weight loss landscape}}
    \end{minipage}
    \caption{\textbf{The ablation studies of WRF4CIR.} (a) Effect of perturbation strength $\gamma$ on test performance; (b), (c), and (d) Visualization of the learning curve, generalization gap, and weight loss landscape under different perturbation strength $\gamma$ on FashionIQ.}
    \Description{The ablation studies of WRF4CIR. (a) Effect of perturbation strength $\gamma$ on test performance; (b), (c), and (d) Visualization of the learning curve, generalization gap, and weight loss landscape under different perturbation strength $\gamma$ on FashionIQ. These results collectively demonstrate that WRF4CIR effectively enhances generalization by flattening the loss landscape without significantly impairing the training process.}
    \label{fig5}
\end{figure*}

\Cref{fig4} illustrates three examples with and without our method. \Cref{fig4}(a) and (b) are drawn from the CIRR dataset, while \Cref{fig4}(c) is from the FashionIQ dataset. The top-5 retrieval results are presented, with the target image highlighted by a green bounding box. In (a) and (b), although the baseline method successfully retrieves the target image within the top-5 results, the ranking of similar samples does not adequately align with the user’s intent. For instance, in (a), the top-1 result contains three hyenas, partially matching the textual description “three hyenas.” However, it fails to capture the primary subject of the reference image (the cow) as well as the relationship between the hyenas and the cow described in the text. In contrast, WRF4CIR correctly ranks the target image as the most similar result and places this highly similar sample in the second position. On the FashionIQ dataset, incorporating WRF enables the model to retrieve the target image within the top-5 results, which is consistent with the observed improvement in recall performance. These results demonstrate that the effectiveness of our method in understanding reference image and text qualitatively, enabling more accurate target image retrieval.\begin{table}[t]
\centering
\captionsetup{position=top}
{\caption{\textbf{Ablation study} towards the weight perturbation strategy of WRF4CIR.}\label{table ablation}}
    
   \resizebox{0.47\textwidth}{!}{
   \begin{tabular}{c cc ccc ccc} 
        \toprule
        \multirow{2}{*}{Method}&\multirow{2}{*}{PD}&\multirow{2}{*}{PS}&\multicolumn{3}{c}{$\textbf{FashionIQ}$}&\multicolumn{3}{c}{$\textbf{CIRR}$}\\
        \cmidrule(lr){4-6} \cmidrule(l){7-9}
        &&&R10&R50&Rmean&R1&R5&R10\\
        \midrule
        $\text{Baseline}$&-&-&54.57&75.02&64.80&55.36&85.16&91.89\\
        $\text{WRF4CIR}_{\mathrm{RWP}}$&-&\checkmark&56.02&76.34&66.18&56.78&85.69&91.86\\
        \cmidrule{1-9}
        \textbf{WRF4CIR}&\checkmark&\checkmark&\textbf{57.35}&\textbf{77.33}&\textbf{67.34}&\textbf{57.85}&\textbf{86.82}&\textbf{92.70}\\
        \bottomrule
   \end{tabular}}
\end{table}
\subsection{Ablation Studies}
In this section, we first analyze the effects of weight perturbation and conduct visualization experiments to evaluate the impact of our method on overfitting in CIR. We then evaluate our method across different CIR mechanisms, data sizes, LoRA ranks, and regularization strategies. Finally, we present an analysis of the computational cost.

\subsubsection{Analysis on weight perturbation.} We first investigate the effects of both the strength and direction of weight perturbations on model performance. The \textit{perturbation strength} $\gamma$ (PS), defined as the ratio between the norm of weight perturbations and that of model parameters, determines the overall intensity of regularization applied to the model. To investigate its impact on retrieval performance, we conducted experiments under different perturbation strengths. As shown in Figure~\ref{fig5}(a), increasing $\gamma$ from zero gradually improves performance on both CIRR and FashionIQ val sets. Experimental results show that WRF4CIR achieves superior performance on both the FashionIQ and CIRR datasets when $\gamma$ is around 0.001. Moreover, the impact of $\gamma$ on model performance typically follows a consistent rise-then-fall trend, with a noticeable improvement observed within the same range. 

Then we investigate the effect of \textit{perturbation direction} (PD) by comparing WRF4CIR with a variant that applies random weight perturbations sampled from $\mathcal{N}(0, 1)$, denoted as $\text{WRF4CIR}_{\mathrm{RWP}}$. The results are shown in \Cref{table ablation}. It can be seen that both the strength and direction of weight perturbations influence model performance. Compared to the baseline introduced in Section~\ref{sec4}, $\text{WRF4CIR}_{\mathrm{RWP}}$ improves FashionIQ Rmean and CIRR R@1 by 1.38\% and 1.42\%, respectively. Further constraining the perturbation direction (WRF4CIR) yields additional gains of 1.16\% and 1.07\% over its RWP counterpart. 
\subsubsection{Visualization.} To further demonstrate the effectiveness of our method, we analyze the \textit{learning curves, generalization gap, and weight loss landscapes} on FashionIQ (\Cref{fig5}(b-d)). As shown in \Cref{fig5}(b-c), increasing the perturbation strength $\gamma$ not only delays the peak performance epoch but also consistently narrows the generalization gap, aligning the test curve more closely with the training trajectory. This improved generalization is further elucidated by the weight loss landscape in \Cref{fig5}(d). We observe that larger $\gamma$ values (up to 0.005) lead to progressively flatter landscapes, which typically signifies enhanced robustness to weight perturbations. Notably, the optimal configuration ($\gamma=0.001$) achieves a superior balance between a flat landscape and low training loss. These results collectively demonstrate that WRF4CIR effectively enhances generalization by flattening the loss landscape without significantly impairing the training process.
\begin{table}[t]
\centering
\captionsetup{position=top}
\caption{\textbf{Effectiveness of our method on existing CIR approaches (Average only).} The best results are shown in bold.}
\label{ablation_existing}
\resizebox{0.85\columnwidth}{!}{
\begin{tabular}{l l c c c}
\toprule
Method & Backbone & R10 & R50 & Rmean \\
\midrule
CLIP4CIR \citep{baldrati2023composed} & ViT-B & 42.87 & 66.77 & 54.82 \\
CLIP4CIR+Ours & ViT-B & \textbf{47.28} & \textbf{69.26} & \textbf{58.27} \\
\midrule
SPRC \citep{bai2023sentence} & ViT-G & 54.92 & 74.97 & 64.85 \\
SPRC+Ours & ViT-G & \textbf{56.67} & \textbf{77.12} & \textbf{66.90} \\
\bottomrule
\end{tabular}
}
\end{table}
\begin{figure*}
    \centering
    \captionsetup{justification=centering}
    \begin{minipage}[t]{0.26\textwidth}
        \includegraphics[width=\textwidth]{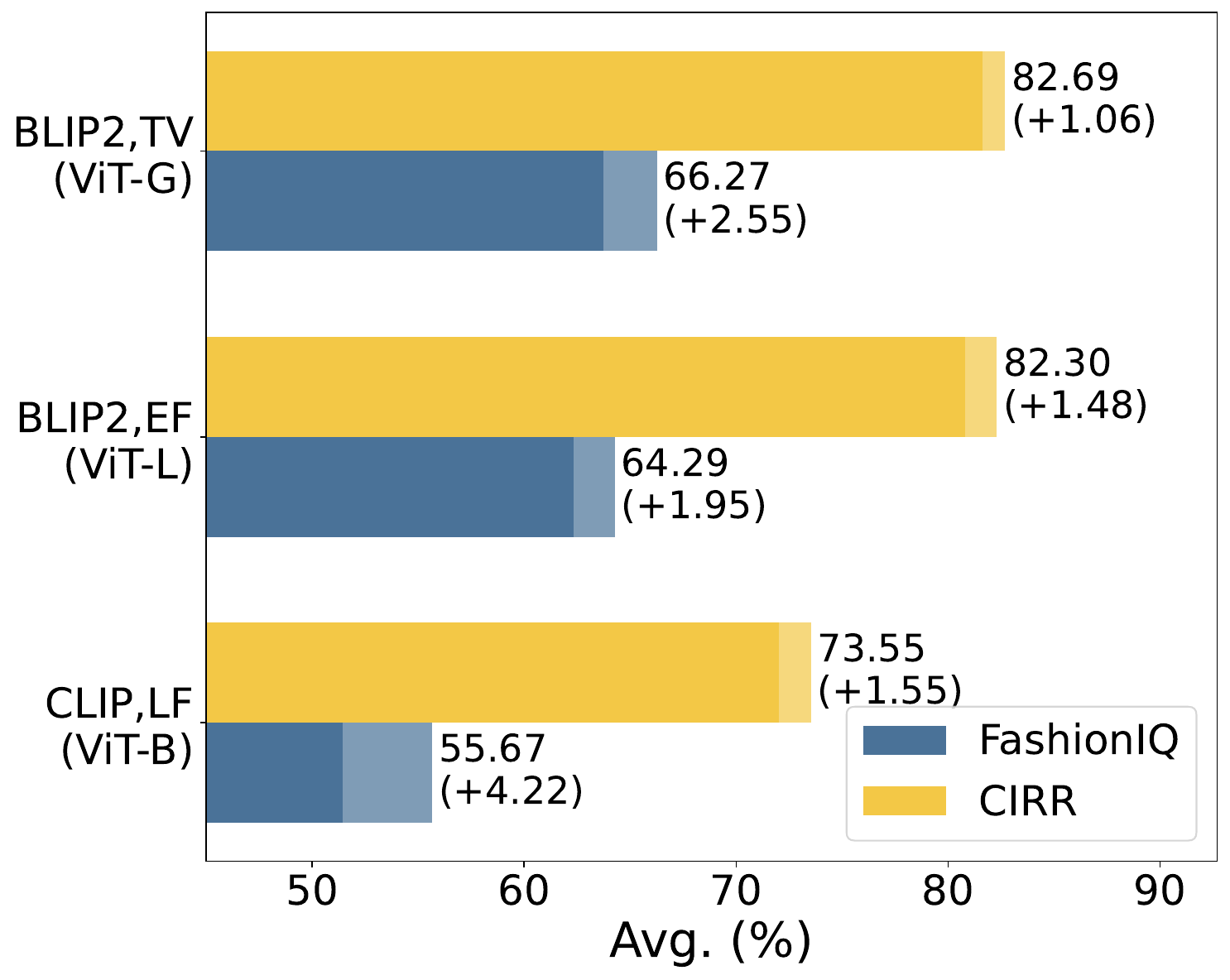}
        \caption*{(a) CIR mechanisms}
    \end{minipage}
    \begin{minipage}[t]{0.23\textwidth}
        \includegraphics[width=\textwidth]{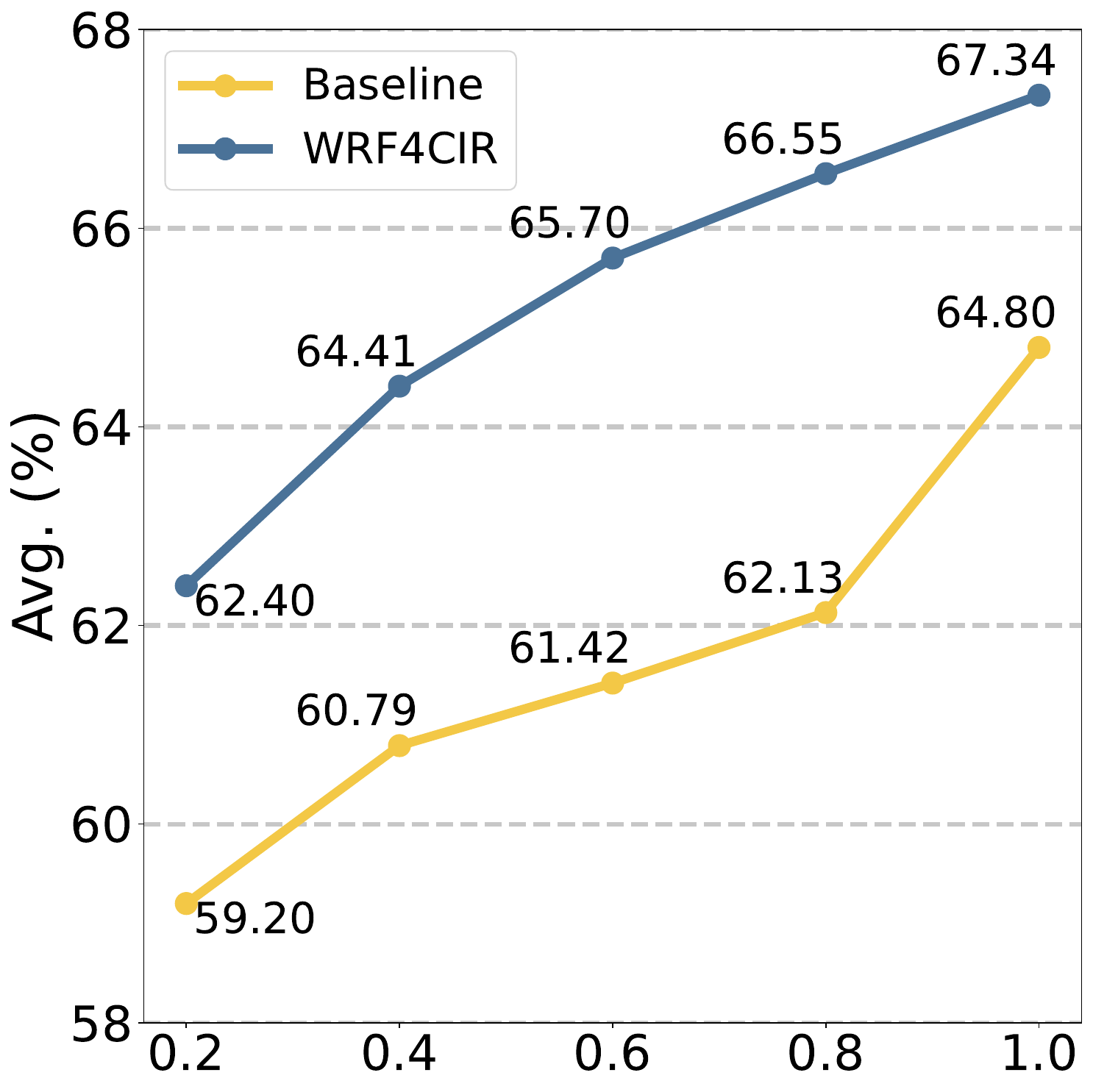}
        \caption*{(b) Data sizes}
    \end{minipage}
    \begin{minipage}[t]{0.23\textwidth}
        \includegraphics[width=\textwidth]{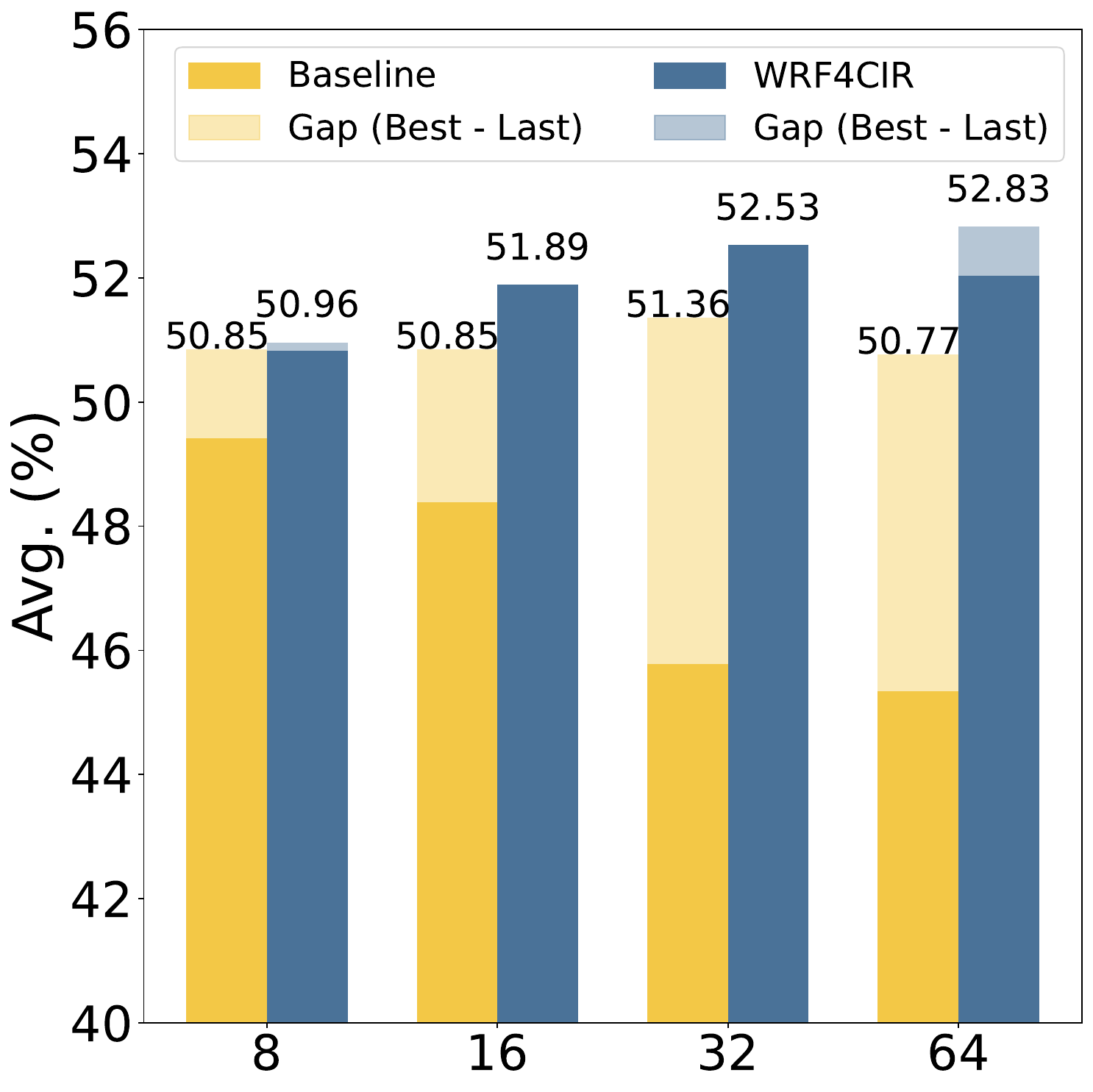}
        \caption*{(c) LoRA ranks}
    \end{minipage}
    \begin{minipage}[t]{0.23\textwidth}
        \includegraphics[width=\textwidth]{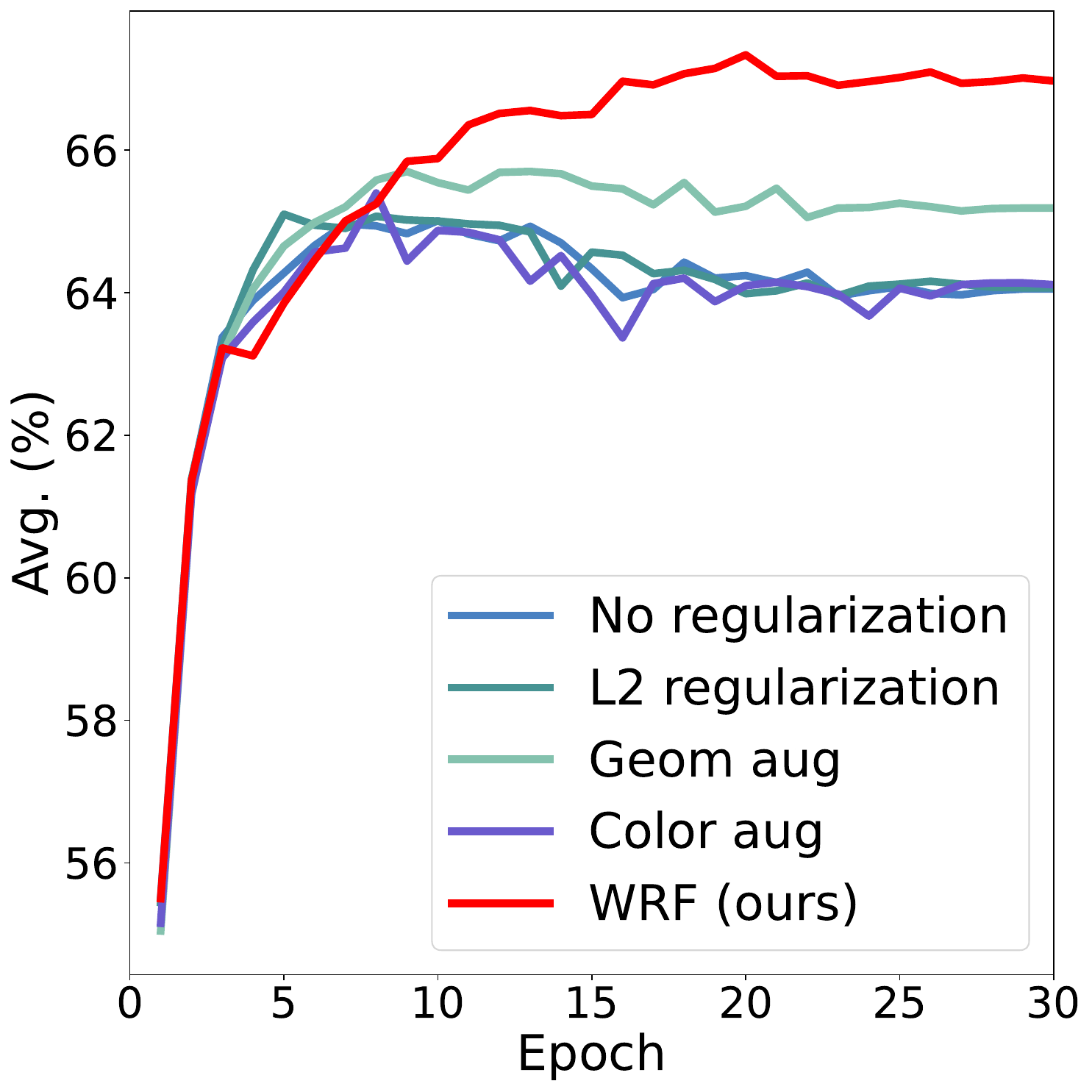}
        \caption*{(d) Regularization strategies}
    \end{minipage}
    \caption{Evaluation of WRF4CIR under different CIR mechanisms, training data sizes, \\LoRA ranks and regularization strategies.}
    \Description{Evaluation of WRF4CIR under different CIR mechanisms, training data sizes, \\LoRA ranks and regularization strategies. These experiments demonstrate that our method achieves substantial improvements across multiple experimental settings.}
    \label{fig6}
\end{figure*}
\subsubsection{Effectiveness in different CIR mechanisms.} CIR mechanisms involve the choice of pre-trained models and fusion strategies, which are crucial for retrieval performance and have been extensively explored in prior work \citep{delmas2022artemis, chen2022composed, baldrati2023composed,bai2023sentence,xu2024set}. To evaluate the effectiveness of weight-regularized fine-tuning across different CIR mechanisms, we conduct experiments using three fusion strategies under two pre-trained models (CLIP and BLIP-2). Specifically, we consider CLIP with late fusion, BLIP-2 with early fusion, and BLIP-2 with text inversion. Implementation details are provided in the supplementary material. As shown in \Cref{fig6}(a), our method results in consistent improvement across different CIR mechanisms on the FashionIQ and CIRR datasets. Our method’s effectiveness is also validated on two baseline approaches, including CLIP4CIR \citep{baldrati2023composed} and SPRC \citep{bai2023sentence}. The experimental results on the FashionIQ dataset are provided in \Cref{ablation_existing}. These results underscore the prevalence of overfitting in VLP-based CIR and confirm our method's effectiveness across various pre-trained models and fusion strategies.
\subsubsection{Analysis on data size.} Then, we evaluate WRF4CIR across different data sizes. Specifically, we divide the FashionIQ training set into different proportions: \{0.2, 0.4, 0.6, 0.8, 1.0\}, while keeping the validation set unchanged, and compare average recall on the FashionIQ val set. As shown in \Cref{fig6}(b), the proposed weight-regularized fine-tuning network is effective across varying data sizes, yielding performance gains of \{+3.2, +3.62, +4.28, +4.42, +2.54\}, respectively. When using only 40\% of the original dataset, WRF4CIR achieves test performance close to full-data standard fine-tuning. These experimental results demonstrate that our approach can effectively fine-tune pre-trained models with limited triplets.
\subsubsection{Effectiveness in parameter-efficient fine-tuning.} To study the effectiveness of WRF4CIR under different fine-tuning strategies, we also consider LoRA-based fine-tuning \citep{hu2022lora} in addition to the standard full-parameter setting commonly used in CIR. Specifically, we adopt CLIP as the backbone and conduct experiments across different ranks on the FashionIQ and CIRR datasets. The FashionIQ result are shown in \Cref{fig6}(c), while additional results for other datasets are provided in the supplementary material. It is observed that WRF4CIR not only achieves superior retrieval accuracy but also significantly narrows the performance gap between the final and peak results. 
\subsubsection{Comparison with different regularization strategies.} We conducted ablation studies to compare WRF (ours) with commonly used regularization strategies, including L2 regularization and data augmentation. Among them, the data augmentation comprises Geom-Aug (e.g., Cutout \citep{devries2017improved}) and Color-Aug (e.g., ColorJitter). As shown in \Cref{fig6}(d), WRF consistently outperforms both L2 regularization and data augmentation methods. Further experimental results and analyses are provided in the supplementary material. These ablation studies confirm that WRF provides better regularization and generalization when fine-tuning pre-trained models.\begin{table}[t]
\centering
\captionsetup{position=top}
\caption{\textbf{Ablation study of performance and training time under different adversarial ratios.}}
\label{computation_performance}
\resizebox{0.8\columnwidth}{!}{
\begin{tabular}{l c c c c c c}
\toprule
Method & Ratio & R10 & R50 & Rmean & Time &\\
\midrule
Baseline&--- &54.57 &75.02 &64.80 &329s/epoch &\\
$\text{WRF4CIR}_{\mathrm{RWP}}$&0 & 56.02&76.34 &66.18 &336s/epoch &\\
\midrule
\multirow{4}{*}{WRF4CIR}&0.25 &56.53 &76.67 &66.60 & 388s/epoch&\\
&0.5 &56.99 &76.67 &66.83 &436s/epoch &\\
&0.75 &57.07 &76.95 &67.01 &490s/epoch &\\
&1.0& 57.35&77.33 &67.34 &573s/epoch& \\

\bottomrule
\end{tabular}
}
\end{table} 
\subsubsection{Computational cost and efficiency.} Overfitting is a prevalent phenomenon in VLP-based CIR, which severely degrades model performance. To address this issue, we introduce a weight-regularized fine-tuning strategy that alleviates overfitting but requires additional computation for generating adversarial perturbations. To balance performance and computational cost, we replace adversarial perturbations with random perturbations in a controlled ratio; the corresponding results are reported in \Cref{computation_performance}. When only random perturbations are applied, our method achieves a 1.38\% performance improvement over the baseline with negligible computational overhead. As the proportion of adversarial perturbations increases, the retrieval accuracy can be further improved. More importantly, our method is applied only during fine-tuning, thereby introducing no additional inference overhead.

\section{Conclusion}
\label{sec:conclu}
In this work, we find that overfitting is a prevalent issue in VLP-based Composed Image Retrieval (CIR), which has become a major bottleneck for further performance improvement. To address this problem, we introduce a Weight-Regularized Fine-tuning network for CIR, termed WRF4CIR. It regularizes the fine-tuning process by introducing adversarial perturbations to the model weights. Extensive experiments demonstrate that our method significantly narrows the generalization gap and achieves substantial improvements over existing methods. 

\bibliographystyle{ACM-Reference-Format}
\bibliography{main}
\end{document}